\documentclass[acmlarge,nonacm]{acmart}
\AtBeginDocument{\settopmatter{printacmref=true}}
\usepackage{epigraph}

\usepackage{natbib}
\usepackage{cleveref}
\usepackage[most]{tcolorbox}
\usepackage{adjustbox}
\usepackage{booktabs}
\usepackage{tabularx}

\usepackage{colortbl}

\AtBeginDocument{%
  }

\copyrightyear{2026}
\acmYear{2026}
\setcopyright{cc}
\setcctype{by}
\acmConference[FAccT '26]{The 2026 ACM Conference on Fairness, Accountability, and Transparency}{June 25--28, 2026}{Montreal, QC, Canada}
\acmBooktitle{The 2026 ACM Conference on Fairness, Accountability, and Transparency (FAccT '26), June 25--28, 2026, Montreal, QC, Canada}
\acmDOI{10.1145/3805689.3806731}
\acmISBN{979-8-4007-2596-8/2026/06}
\acmConference[ACM FAccT 2026]{ACM Conference on Fairness, Accountability, and Transparency 2026}{June 25--28, 2026}{Montreal, QC, Canada}
\begin{document}

\title[Normative Common Ground Replication (NormCoRe)]{Normative Common Ground Replication (NormCoRe): Replication-by-Translation for Studying Norms in Multi-Agent AI}


\author{Luca Deck}
\authornote{These authors contributed equally to this work.}
\email{luca.deck@uni-bayreuth.de}
\orcid{0000-0003-3773-2769}
\affiliation{
    \institution{University of Bayreuth \& Fraunhofer FIT}
    \city{Bayreuth}
    \country{Germany}
}

\author{Simeon Allmendinger}
\authornotemark[1]
\email{simeon.allmendinger@uni-bayreuth.de}
\orcid{0009-0005-8741-7734}
\affiliation{
    \institution{University of Bayreuth \& Fraunhofer FIT}
    \city{Munich}
    \country{Germany}
}

\author{Lucas Müller}
\email{lucas.c.mueller@gmail.com}
\orcid{0009-0000-0405-8427}
\affiliation{
    \institution{University of Bayreuth}
    \city{Bayreuth}
    \country{Germany}
}

\author{Niklas Kühl}
\email{kuehl@uni-bayreuth.de}
\orcid{0000-0001-6750-0876}
\affiliation{
    \institution{University of Bayreuth \& Fraunhofer FIT}
    \city{Bayreuth}
    \country{Germany}
}


\begin{abstract}
In the late 2010s, the fashion trend \textit{NormCore} framed sameness as a signal of belonging, illustrating how norms emerge through collective coordination.
Today, similar forms of normative coordination can be observed in systems based on Multi-agent Artificial Intelligence (MAAI), as AI-based agents deliberate, negotiate, and converge on shared decisions in fairness-sensitive domains.
Yet, existing empirical approaches often treat norms as targets for alignment or replication, implicitly assuming equivalence between human subjects and AI agents and leaving collective normative dynamics insufficiently examined.
To address this gap, we propose \textit{Normative Common Ground Replication} (\textit{NormCoRe}), a novel methodological framework to systematically translate the design of human subject experiments into MAAI environments.
Building on behavioral science, replication research, and state-of-the-art MAAI architectures, \textit{NormCoRe} maps the structural layers of human subject studies onto the design of AI agent studies, enabling systematic documentation of study design and analysis of norms in MAAI.
We demonstrate the utility of \textit{NormCoRe} by replicating a seminal experimental study on distributive justice, in which participants negotiate fairness principles under a ``veil of ignorance''.
We show that normative judgments in AI agent studies can differ from human baselines and are sensitive to the choice of the foundation model and the language used to instantiate agent personas.
Our work provides a principled pathway for analyzing norms in MAAI and helps to guide, reflect, and document design choices whenever AI agents are used to automate or support tasks formerly carried out by humans.
\end{abstract}





\maketitle
\vspace{50pt}
\section{Introduction}
\textit{``Once upon a time people were born into communities and had to find their individuality.
Today people are born individuals and have to find their communities.''}
\begin{flushright}
--- K-HOLE, inventors of the term ``Normcore''
\end{flushright}

In the late 2010s, the fashion trend \emph{NormCore} embraced deliberate sameness: conventional clothing as a signal of belonging rather than a mark of distinction.
What appeared aesthetic was fundamentally normative; an emergent agreement about what is considered appropriate within a group.
Such norms arise not from isolated individuals, but from collective coordination.
Today, similar forms of normative coordination are increasingly relevant in systems based on Multi-agent AI (MAAI).
In contrast to individual AI agents, in MAAI~\citep{Li_2024_MAAI_Paradigm, allmendinger2025multi} AI-based agents can deliberate, negotiate, and converge on shared decisions.
In doing so, they explicitly or implicitly exhibit social and ethical norms~\citep{Cui2025}, in particular as MAAI systems are already being developed in domains governed by fairness and other norms, such as resource allocation~\citep{zhangEfficientLLMGrounding2025} or autonomous driving~\citep{taghavifarBehaviorallyAwareMultiAgentRL2025}.
If we examine this field from a more traditional perspective, experimental research on social and ethical norms is deeply rooted in philosophy, psychology, and game theory, with well-established methods~\citep{bicchieri2005grammar,hechterSocialNorms2001, lewis1969convention}.
For example, preferences regarding wealth distribution have been extensively studied through game-theoretical group experiments~\citep{frohlichChoosingJusticeExperimental1992, durantePreferencesRedistributionPerception2014}.
However, what these studies all have in common is that they are examining norms among \textit{humans}.

\begin{figure*}
    \centering
    \includegraphics[width=\linewidth]{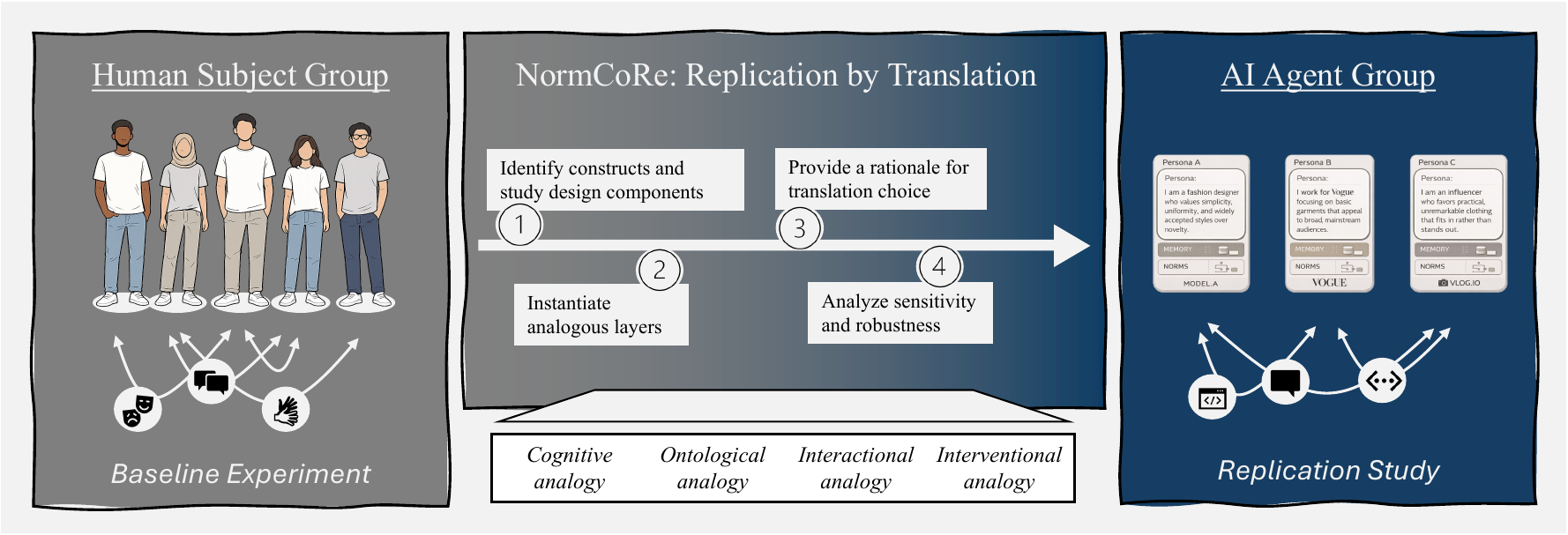}
    \caption{From human groups to multi-agent AI: \emph{NormCoRe} conceptualizes replication as a translation problem, mapping human subject studies to AI agent studies to study how collective normative judgments---such as fairness---emerge and differ across populations.}
    \label{fig:graphical_abstract}
\end{figure*}

Conversely, research on AI agents---particularly AI agent groups---has only sparsely engaged with the methodological foundations of studying social and ethical norms.
At the same time, recent research is already proposing to replace or supplement human subjects with AI agents~\citep{bailCanGenerativeAI2024, xuAISocialScience2024}, replicating existing human subject studies using AI agents~\citep{Cui2025}, or attempting to align AI agents with normative principles~\citep{weidingerUsingVeilIgnorance2023a}.
Without systematically acknowledging the fundamental differences between human subjects and AI agents, this obscures a critical question for fairness and accountability: how do collective normative judgments emerge, stabilize, and differ when decision-making is delegated to MAAI rather than human groups?
Regardless of the underlying goals of AI agent studies, the complexities arising from translating human subject studies to MAAI necessitate a sound methodological foundation that is currently lacking---particularly when social and ethical norms are the focus of such studies.
For example, when MAAI systems are tasked with making implicit or explicit decisions about resource allocations, the sheer number of degrees of freedom in the design of the MAAI system, ranging from the selection of a foundation model (e.g., Large Language Model (LLM)) to the design of task-specific workflows, impedes a thorough evaluation of design choices.

Against this backdrop, we propose Normative Common Ground Replication \textit{NormCoRe}\footnote{The method does not coincidentally bear the same name as the fashion phenomenon described above.}: a novel methodological framework for translating human subject group studies into MAAI environments, enabling researchers to \textit{systematically} investigate social norms in AI agent studies (see~\Cref{fig:graphical_abstract}).
Building on established principles from behavioral science and MAAI research, NormCoRe maps the analogous layers of human subject studies onto the design of AI agent studies with MAAI.
NormCoRe allows researchers to systematically select and document the configuration of AI agent studies.

We demonstrate the utility of NormCoRe by replicating \citet{frohlichChoosingJusticeExperimental1992}'s influential study on distributive justice in an MAAI setting, the complete code including all prompts used is available on Github~\citep{deck2026normcore_software}.
The selected baseline study serves as a perfect showcase, as it conceptualizes a complex but well-known norm (e.g., fairness) as a group-level, normative judgment reached through dynamic deliberation and includes all relevant layers for replication.
In the original experiment, \citet{frohlichChoosingJusticeExperimental1992} operationalized John Rawls' veil of ignorance''~\citep{Rawls_1971_theory_justice} within a controlled laboratory setting.
Participants were unaware of their assigned income class---simulating the ``veil of ignorance''---and were tasked with reaching a consensus on a distributive justice principle that would determine their payoff in a hypothetical society.
By combining individual normative deliberation, goal-based optimization, and dynamic consensus finding, this experiment serves as an ideal testbed for demonstrating how NormCoRe organizes complex design choices and facilitates precise reporting.
Instantiating NormCoRe with \citet{frohlichChoosingJusticeExperimental1992}'s study, we show that the choice of the LLM and the language of the persona description have a significant influence on fairness judgments in AI agent studies.
Also, we find that while both populations favor the same principle (maximizing overall income while ensuring a floor constraint for the worst-off), MAAI groups demonstrate a substantially higher concentration on this principle than human groups.

This work makes three contributions to the rigorous design and analysis of MAAI:
\begin{itemize}
    \item We establish a novel \textit{replication-by-translation} perspective on replication studies with AI agents that explicitly accounts for the fundamental differences between AI agents and human subjects (\Cref{sec:background}).
    \item Based on this perspective, we introduce NormCoRe as a methodological framework for the systematic replication of human subject studies in MAAI settings, providing a lens through which researchers can document and analyze social and ethical norms in AI agents (\Cref{sec:method}).\footnote{Our codebase repository is available under \url{https://github.com/Lucas-Mueller/Normative_Common_Ground_Replication_NormCoRe}}
    % and compare them to the outcomes of human subject studies 
    \item By employing NormCoRe to a seminal baseline study on distributive justice, we demonstrate the usefulness of NormCoRe and empirically show that norms in MAAI not only differ from human baselines but are also sensitive to study design decisions~(\Cref{sec:results}).
\end{itemize}

We discuss the implications of our study and broaden the discourse on the purpose and open challenges of AI agent studies in \Cref{sec:discussion}, paving the way for future research on norms in MAAI.
It is reasonable to expect that the implementation of MAAI in studies and industry will happen one way or another. However, to judge whether this is a positive or a reprehensible development---or under which conditions MAAI systems and studies are actually beneficial---we need to better understand the structure, dynamics, and impact of MAAI systems in the first place.
Our work raises critical normative questions regarding the future of agentic automation beyond distributive justice and cautions designers to make conscious, evidence-based choices in both laboratory experiments and real-world applications.
Whenever tasks formerly carried out by (groups of) humans are to be automated or supplemented by groups of AI agents, NormCore helps to guide, reflect, and document design choices and to study potential impact of including AI agents.
\section{Background} \label{sec:background}
This section reviews prior work on replicability as a methodological principle~(\Cref{background:replicability}) and its application to AI-based replication of human subject studies.
We then identify key assumptions and challenges in existing approaches, particularly the treatment of human--AI replication as equivalence rather than translation~(\Cref{background:AIagents}), which motivates the need for a novel methodology~(\Cref{background:translation}).

\subsection{Importance of Replicability for the Scientific Method} \label{background:replicability}
Replicability constitutes a fundamental pillar of the scientific method. 
In principle, the transparent documentation of research methodologies enables independent verification and replication of empirical findings by the broader academic community.
In that sense, replication can serve two functions: authentication of original findings and boundary testing to understand generalizability~\citep{zwaan2018making}. 
Consequently, two replication approaches are distinguished: \emph{direct replication}, which repeats an experiment with minimal to no changes to authenticate original findings, and \emph{conceptual replication}, which tests the same hypothesis with different methods, stimuli, or populations, aiming to probe whether findings generalize to new conditions~\citep{zwaan2018making,koehler2021play}. 
Although sometimes treated as synonyms, a critical distinction exists between reproducibility, where existing data is re-analyzed with the same methods, and replicability, where experiments are repeated, resulting in new data~\citep{Patil.2016}.

\subsection{Replicability of Human Subject Studies with AI Agents} \label{background:AIagents}
Meanwhile, a growing body of research across disciplines is conceptually replicating human experiments with LLM-based AI systems~\citep{lee2024large,hagendorff2023human,akata2025playing,mittelstadt2024large,grizzard2025chatgpt,binz2025foundation}.
These efforts pursue two distinct objectives.
First, replication is employed to assess whether AI agents can serve as a valid proxy for human participants in experimental research~\citep{Yeykelis2024,Cui2025}.
Second, replication is used to compare humans and AI as a lens for better understanding the psychological, behavioral, and normative properties of AI systems~\citep{Leng2024}.

Work pursuing the first objective typically evaluates replication success in terms of statistical similarity between human and AI-generated responses.
For example, \citet{Yeykelis2024} assessed the replicability of consumer behavior research by replicating 133 results from 45 studies published in the Journal of Marketing, using LLM-based AI personas programmed to match the original participant demographics.
They report that 76\% of main effects and 68\% of interaction effects could be reproduced.
Similarly, in Psychology and Management Science, \citet{Cui2025} replicated 156 experiments published over the past decade and found that in 73\% and 81\% of main effects, respectively, were replicated, with interaction effects rates ranging between 46\% and 63\%.

In contrast, work aligned with the second objective treats replication outcomes not as validation but as a diagnostic tool for understanding how and where AI behavior diverges from human cognition.
In behavioral economics, \citet{Leng2024} replicated canonical experiments on prospect theory, framing, and mental accounting (e.g.,~\citep{KahnemanTversky1979,Thaler1985}).
While LLMs partially reproduce human mental accounting behavior, they appear substantially more rational, exhibiting weak framing effects and limited transaction utility.
Crucially, these behavioral patterns vary systematically with design choices such as the language of the prompt, with Spanish and French prompts showing more human-like loss aversion than English and Chinese prompts~\citep{Leng2024}.

A related line of work employs replication to assess whether MAAI aligns with human moral judgments, implicitly treating human consensus as a normative benchmark.
Within this approach, moral soundness is operationalized as statistical similarity between human and AI responses to ethically charged dilemmas.
A prominent example is the replication of the Moral Machine experiment, in which human participants were asked to resolve trolley-problem-style dilemmas and their aggregate judgments were taken as indicative of moral preferences~\citep{Awad.2018}.
When this experiment is applied to LLM-based agents, replication results show that moral judgments vary substantially depending on model architecture and training regime~\citep{Takemoto2024}.
Such approaches also gain traction in the industry, as evidenced by Anthropic's creation and use of the GlobalOpinionQA dataset, which contains 2556 questions and human answers on ethics and current events from the World Values and Pew Global Attitudes surveys~\citep{Durmus.2023}.
Crucially, Anthropic evaluates their LLM against the normative target that the model should reflect a country's specific distribution of opinions when prompted---implying, e.g., that a model prompted in Russian ought to adopt a distinctively ``Russian'' moral perspective.

The implicit assumption underlying this stream of research is that convergence with human consensus constitutes ethical adequacy~\citep{lacroixMoralDilemmasMoral2022}.
From a philosophical perspective, this assumption mirrors the naturalistic fallacy, which cautions against ``oughts'' from descriptive ``is'' statements about observed human behavior~\citep{Moore.1903}.
A more fundamental limitation of this literature concerns the conception of the experimental subject itself.
Many replication studies implicitly treat LLM-based agents as functional substitutes for human participants, evaluating success primarily in terms of behavioral or statistical similarity~\citep{lee2024large,mittelstadt2024large,binz2025foundation}.
However, an LLM is fundamentally different from a human being with physical embodiment, lived experience, and moral agency, and a prompt-based persona is not ontologically equivalent to a human subject.
By overlooking these differences, replication is often framed as a direct transfer rather than as a translation between fundamentally different kinds of subjects, whose cognitive substrates, experience, and agency differ in principled ways~\citep{hagendorff2023human,akata2025playing,grizzard2025chatgpt}.
These risks obscure how observed similarities or divergences in normative outcomes are shaped by design choices and structural differences rather than genuine equivalence.

\subsection{Replication-by-Translation in AI Agent Studies} \label{background:translation}
The fundamental ontological differences between human subjects and AI agents imply that replication across populations cannot be accommodated through a straightforward transfer of study design.
Instead, studies that replicate human experiments with LLM-based agents face a crucial methodological challenge: each replication necessarily involves translating human experimental constructs into AI-compatible designs.
This translation introduces degrees of freedom at multiple levels of the AI system, including the choice of foundation model encoding background knowledge, the design of personas and prompts that instantiate subject-like behavior, and the orchestration protocols that structure group interaction.
Such translation choices are rarely documented systematically, yet they fundamentally condition the outcomes being compared, as demonstrated in recent research.
For example, slight variations~\citep{sclar2024prompt}, formatting changes~\citep{voronov.2024}, and framing~\citep{brucks2025prompt} in prompt design can alter the performance of an LLM significantly. 
Similarly, in multi-agent settings, modifications to deliberation protocols, including speaking order~\citep{baltaji.2024}, debate termination timing, and adversarial structure play a decisive role in determining collective outcomes~\citep{liang2024debate}.

Yet, the replication studies reviewed above typically treat these design choices as implementation details rather than methodological choices.
This impedes rigorous interpretation of study results. For example, when human and AI populations yield different outcomes, it cannot be determined whether the differences reflect genuine divergence or a result of arbitrary translation choices.
Addressing this rigor requires treating replication across populations as an explicit translation problem---one that renders translation decisions visible at each layer of the experiment.
Only then can observed differences be attributed to specific sources rather than confounded across the translation process.
To facilitate the standardization of replication practices in line with broader scientific goals~\citep{zwaan2018making}, we propose \textit{NormCoRe} as a novel methodological framework for AI agent-based replication studies.
\section{Normative Common Ground Replication (NormCoRe)} \label{sec:method}

NormCoRe is a systematic method for replicating \emph{human subject studies} with \emph{AI agent studies} in order to (empirically) investigate the normative common ground between human and AI groups.
NormCoRe adapts established replication logic from the social sciences to cross-population settings involving fundamentally different kinds of groups.
Rather than assuming equivalence between human subjects and AI agents, NormCoRe explicitly treats replication as a translation problem to systematically map the constructs, interventions, and interactions of human-subject studies into an analogous study with AI agents, while minimizing threats to validity.
The central goal of NormCoRe is not to determine whether AI agents behave ``like humans,'' but to identify where normative (group) judgments converge or diverge, and to attribute such differences to replication choices.
This enables a disentangled investigation of questions such as whether fairness judgments in MAAI systems are sensitive to model choice, persona, or memory design, and how orchestration and coordination mechanisms shape collective decision-making trajectories.

\begin{figure*}
    \centering
    \includegraphics[width=\linewidth]{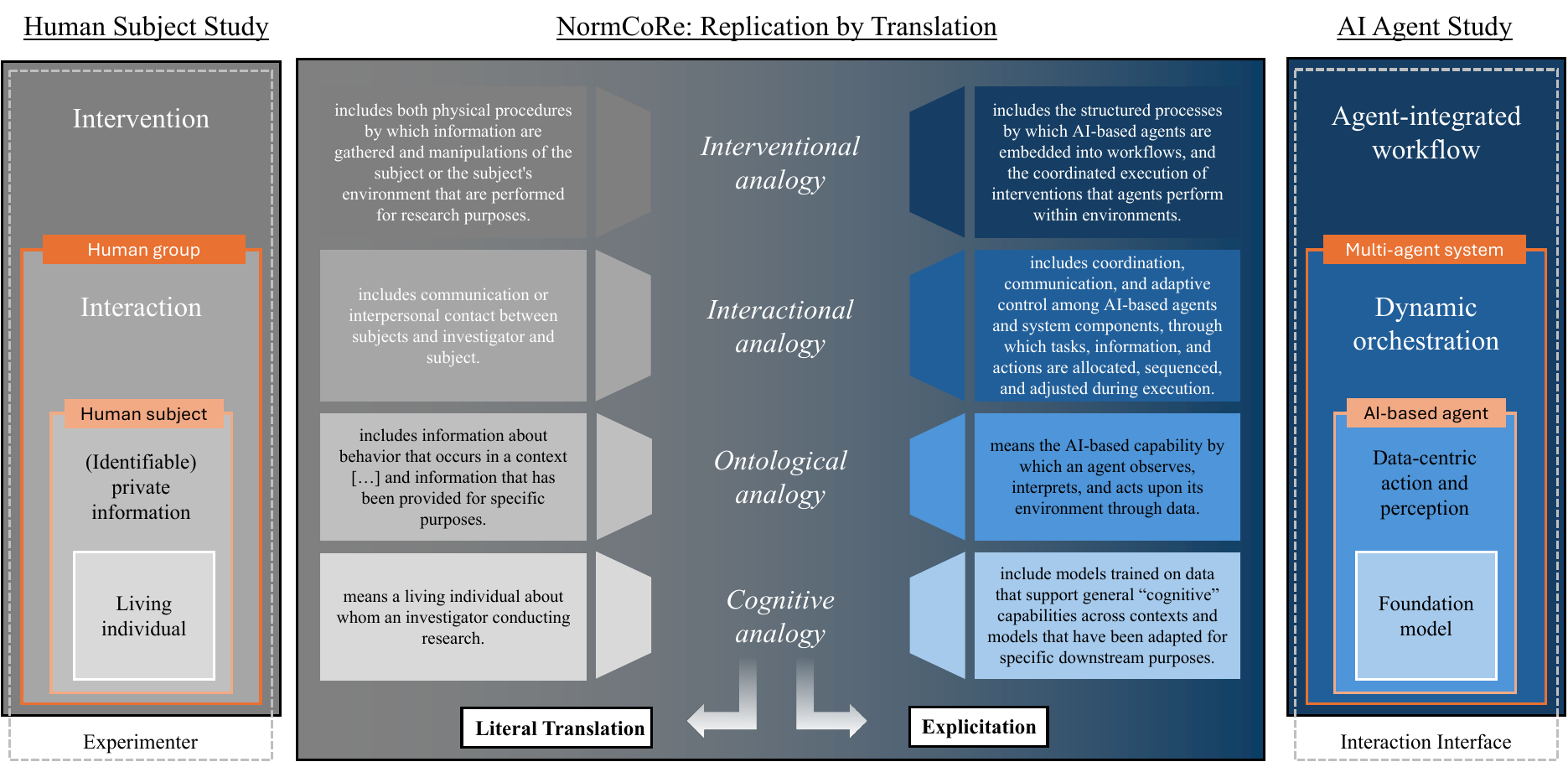}
    \caption{The four translation layers illustrating the necessary analogies between individual layered components of human subject studies and AI agent studies.
    Some components may be translated ``literally'', e.g., when the study sequence can be fully adopted.
    Other components may require ``explicitation'', e.g., when AI agents participate in a discussion in fixed turns.}
    \label{fig:normcore}
\end{figure*}

\subsection{NormCoRe as Replication-by-Translation}
In contrast to direct replication~\citep{tsang1999replication} within a single population, cross-group replication between humans and AI agents necessarily introduces conceptual degrees of freedom arising from differences in four translation layers: \textit{cognition analogy, ontological analogy, interactional analogy} and \textit{interventional analogy}.
We distinguish two ideal-typical translation choices as methodological parameters to implicitly acknowledge underlying differences:
(i) \emph{literal translation for direct replication}, which aims to preserve surface-level experimental features (e.g., task structure, payoff matrices, information availability), 
(ii) \emph{explicitation for analogous replication}, which makes implicit assumptions in human studies explicit and operationalizes them as design choices in AI agent studies (e.g., decision rules, memory structures).

\subsection{Layered Translation Structure in NormCoRe}

NormCoRe operationalizes replication between human subject studies and AI agent studies through a layer-by-layer translation method illustrated in \Cref{fig:normcore}.
Rather than first defining human subject and AI agent studies independently, NormCoRe aligns them at the four corresponding layers of abstraction, following the nested structure of the U.S. Common Rule for human subject studies (45 CFR §46.102~\citep{cfr45_46_102}) in combination with the layered architecture of multi-agent AI~\citep{allmendinger2025multi}.
This approach makes the translation challenge explicit at each layer, allowing normative outcomes to be interpreted relative to specific sources of variation and reducing the risk of what we term \emph{translation hacking}, i.e., the selective tuning of translation-layer design choices to obtain desired outcomes.

\textbf{Layer 1: Living Individual $\rightarrow$ Foundation Models}.
At the most fundamental layer, human subjects are defined as \emph{living individuals} about whom information is obtained through intervention or interaction~\citep{cfr45_46_102}.
Normative outcomes in human studies are already shaped at this level by background knowledge, lived experience, and informational asymmetries. 
These factors are not themselves experimental manipulations, yet they condition all downstream perception, deliberation, and behavior.
In AI agent studies, the structurally corresponding layer is the \emph{foundation model} substrate.
Foundation models encode large-scale statistical regularities derived from pretraining data and thereby constitute the epistemic and normative background against which all agent behavior unfolds~\citep{bommasani2021opportunities}.
While private information in human studies is generated within the experiment, foundation models represent a pre-experimental informational endowment that cannot be directly manipulated during execution but strongly conditions normative judgments.
NormCoRe treats this correspondence as a cognitive analogy.
The goal is not to equate living individuals with foundation models, but to acknowledge that both possess a base layer of informational constraint that shapes what kinds of cognitive heuristics or normative judgments are even expressible.
Translation choices at this layer (e.g., model family, pretraining corpus) introduce degrees of freedom that must be documented to ensure interpretability of replication results.

\textbf{Layer 2: Human Subject $\rightarrow$ AI-based Agent}.
Building on this informational substrate, the second layer concerns the \emph{subject} itself.
Human subjects are grounded in the notion of (identifiable) private information that is used, studied, analyzed, or generated within the research context~\citep{cfr45_46_102}.
This layer establishes the epistemic basis of the study: what is known about the subject, how that knowledge is obtained, and under which contextual constraints.
This definition carries implicit assumptions about agency, perception, memory, and the capacity to form normative judgments.
In AI agent studies, the corresponding layer is the agent's \emph{data-centric perception–action capability}: the mechanisms through which an agent observes its environment, interprets inputs, and generates actions~\citep{allmendinger2025multi}.
This includes prompt structures, persona specifications, memory mechanisms, and decision heuristics that instantiate agency-like behavior in computational form.
NormCoRe frames this mapping as an \emph{ontological analogy}.
Human agency is not replicated, but functionally translated into computational capacities that allow agents to participate meaningfully in normative tasks.
Explicit design choices at this layer---such as whether agents possess persistent memory or adopt stable personas---directly affect normative outcomes and must therefore be treated as core elements of the replication design rather than as mere implementation details.

\textbf{Layer 3: Human Group $\rightarrow$ Multi-agent System}.
The third layer concerns \emph{interaction}.
In human groups, interaction encompasses communication or interpersonal contact \textit{between} subjects and investigators, as well as \textit{among} subjects.
This layer structures deliberation, persuasion, power dynamics, and social learning, all of which are central to the emergence of shared norms such as fairness~\citep{gelfand2024norm}.
In AI agent studies, the corresponding layer is \emph{dynamic orchestration} creating a Multi-agent system.
This includes message passing protocols, turn-taking rules, negotiation mechanisms, and adaptive control processes that determine how agents exchange information and influence one another over time~\citep{allmendinger2025multi}.
NormCoRe treats this correspondence as an interactional analogy.
Translation at this layer often requires explicitation: assumptions that are often implicit in human interaction (e.g., (equal) speaking rights, shared understanding of rules) must be made explicit as protocol constraints or coordination mechanisms in AI agent studies.

\textbf{Layer 4: Intervention $\rightarrow$ Agent-integrated Workflow}.
At the highest layer, human subject studies involve \emph{interventions}, defined as physical procedures or informational manipulations performed for research purposes~\citep{cfr45_46_102}.
Interventions include task framing, incentive structures, timing, and constraints that are intentionally varied to study causal effects.
In AI agent studies, the corresponding layer is the \emph{agent-integrated workflow}~\citep{allmendinger2025multi}.
This includes the structured processes by which agents are embedded into experimental workflows, the sequencing of tasks, role assignments, and the coordinated execution of actions within an environment.
This mapping constitutes an interventional analogy.
NormCoRe emphasizes that such translations are not neutral: different design choices can systematically shape normative outcomes~\citep{simson2024one}. 
Accordingly, intervention-level explicitation choices must be explicitly justified and reported.

In addition to the formally specified translation layers, the role of the experimenter should be explicitly acknowledged as an ``unknown influence''.
In human subject studies, experimenters inevitably shape outcomes through subtle cues, framing choices, timing, and interaction styles---often unintentionally and outside the scope of formal interventions.
In AI agent studies, analogous influences arise through interaction interfaces.
Making this influence explicit does not eliminate it, but improves interpretability by preventing ungrounded attribution of normative differences solely to agents or models, when they may partly reflect researcher-induced artifacts.
To operationalize this concern and to ensure methodological interpretability, we summarize the NormCoRe method in the following Box.

\begin{tcolorbox}[
    colback=gray!10,
    colframe=gray!60,
    title=\textbf{NormCoRe Method},
    fonttitle=\bfseries,
    sharp corners,
    boxrule=0.8pt,
]
    We propose a method for the interpretable replication of human subject studies with AI agent studies by establishing \emph{disentangled, layer-specific translation analogies}, rather than assuming global equivalence.
    Replication success is evaluated in terms of \emph{explanatory alignment}, not behavioral identity.

\medskip

For each NormCoRe translation exercise, researchers must:

\begin{enumerate}
    \item[\textit{Step 1}] \textbf{Identify constructs \& study design components:}  
    Specify the theoretically relevant constructs in the original human subject study together with the concrete study design components through which these constructs are operationalized (e.g., task framing, information structure, incentive schemes, interaction rules, and experimenter involvement).

    \item[\textit{Step 2}] \textbf{Instantiate analogous layers:}  
    Instantiate the analogous MAAI component(s) of the corresponding NormCoRe layer (e.g., foundation model choice for the cognitive layer, persona prompting and memory for the ontological layer, orchestration protocols for the interactional layer).

    \item[\textit{Step 3}] \textbf{Provide a rationale for translation choices:}  
    Justify whether the layer mapping constitutes:
    \begin{itemize}
        \item \emph{Literal translation} (for exact/direct replication), or
        \item \emph{Explicitation} (for analogous replication), and specify the replication type implied by the design choice (e.g., constructive, incremental, quasirandom, or comprehensive~\citep{koehler2021play}).
    \end{itemize}

    \item[\textit{Step 4}] \textbf{Analyze sensitivity \& robustness:}  
    Analyze the sensitivity of normative outcomes to layer-specific translation choices using established statistical criteria for replication, non-replication, and partial replication evidence~\citep{douglas2021design}.
\end{enumerate}

\end{tcolorbox}
%%%%%%%%%%%%%%%%%%%%%%%%
%% Experimental Study %%
%%%%%%%%%%%%%%%%%%%%%%%%
\section{Experimental Study: Applying NormCoRe for Fairness Principles}
\label{sec:results}
With the methodological framework in place, we illustrate its application for a representative norm that has both a social and ethical dimension as well as a crucial downstream impact in MAAI systems: fairness.
Fairness is a norm deeply embedded in social science and philosophy, guiding perceptions, legal frameworks, economic interactions, and the design of AI systems~\citep{binnsFairnessMachineLearning2018, mulliganThisThingCalled2019}.
While its meaning varies across disciplines, contexts, and cultures, several simplifying approaches have been proposed to study fairness principles in experiments with human subjects.
One seminal experiment is \citet{frohlichChoosingJusticeExperimental1992} who applied John Rawl's popular thought experiment of the ``veil of ignorance''~\citep{Rawls_1971_theory_justice} to study preferences for four predefined fairness principles representing different schools of thought in political philosophy.
The study has been frequently cited and serves as a perfect showcase for the utility of NormCoRe, as it offers value-laden tradeoffs and disagreement between individuals, and includes all relevant layers for replication and sufficiently complex normative interactions to study the dynamics of MAAI (e.g., as opposed to the Ultimatum Game~\citep{Thaler_1988_Ultimatum} with anonymous and transactional interactions).

The following section describes the original study (\Cref{results:baseline}) and instantiates the NormCoRe framework by translating the original study to an MAAI setting~\citep{allmendinger2025multi} to study individual judgments and group dynamics of MAAI systems in the context of fairness principles~(\Cref{results:translation}).
This instantiation is meant to validate the methodological framework in a concrete application and demonstrates the importance of design choices in AI agent studies~(\Cref{results:results}).
%%%%%%%%%%%%%%
%% Baseline %%
%%%%%%%%%%%%%%

\begin{table}[t]
\centering
\caption{Distribution of fairness principle agreements in the human baseline and MAAI replication, including baseline alignment and sensitivity to translation-layer design choices (foundation model and persona language).}
\small  % Makes font slightly smaller to fit width without complex resizing
\setlength{\tabcolsep}{3pt} % Platz zwischen Spalten minimieren
\begin{tabularx}{0.9\textwidth}{| l | c c | >{\centering\arraybackslash}X >{\centering\arraybackslash}X | c c c |}\hline
 & \multicolumn{2}{c|}{Baseline Alignment}
 & \multicolumn{5}{c|}{Translation Sensitivity} \\
\cline{2-8}
 & \multicolumn{2}{c|}{Human--MAAI}
 & \multicolumn{2}{c|}{Cognitive Layer}
 & \multicolumn{3}{c|}{Ontological Layer} \\
\cline{2-8}  % A standard line from column 2 to 8
\textbf{Fairness Principle}
& \textbf{Baseline}
& \textbf{MAAI}
& \textbf{Chinese LLM \ Ecosystem}
& \textbf{U.S. LLM \ Ecosystem}
& \textbf{English}
& \textbf{Mandarin}
& \textbf{Spanish} \\
\hline
Max.\ Floor & 1 & 0 & 14 & 4 & 0 & 1 & 0 \\
Max.\ Avg. Income & 1 & 1 & 0 & 2 & 0 & 0 & 0 \\
Max.\ Avg. + Floor & 23 & 29 & 15 & 21 & 30 & 27 & 17 \\
Max.\ Avg. + Range & 2 & 0 & 0 & 0 & 0 & 0 & 2 \\
No Agreement & 7 & 3 & 4 & 6 & 4 & 6 & 15 \\
\hline
\textbf{Total} & \textbf{34} & \textbf{33} & \textbf{33} & \textbf{33} & \textbf{34} & \textbf{34} & \textbf{34} \\
\hline
\end{tabularx}
\label{tab:merged_results}
\end{table}
\subsection{Baseline Human Subject Study by Frohlich and Oppenheimer} \label{results:baseline}
\citet{frohlichChoosingJusticeExperimental1992} conducted their experiment with 34 groups of five university students, comprising an individual and a group phase.
In the individual phase, participants received an introduction to four fairness principles in the context of income distribution:
(P1) maximizing the income of the worst-off individual;
(P2) maximizing average (and thus total) income;
(P3) maximizing average income subject to a guaranteed minimum income; and
(P4) maximizing average income subject to a cap on income inequality.
Participants first ranked the principles from most to least preferred and reported their confidence on a five-point Likert scale.
They were then shown four alternative income distributions across five income classes with known probabilities representing a ``probabilistic veil of ignorance'' (5\%, 10\%, 50\%, 25\%, and 10\%) and informed which distribution corresponded to each principle.
After additional instruction and a comprehension test, participants ranked the principles again.

Next, participants completed four payoff-relevant practice rounds.
In each round, they selected a principle, after which a corresponding distribution was implemented via a random draw assigning them to one of five income classes, essentially ``lifting'' the ``veil of ignorance''.
While class probabilities were fixed, participants were unaware of their exact values. 
Realized payoffs, as well as counterfactual payoffs under alternative principles, were revealed and paid immediately at a 1:\$10,000 conversion rate.
The individual phase concluded with a third ranking and confidence assessment.
In the group phase, each five-person group deliberated to reach a unanimous agreement on a single principle.
Prior to the discussion, participants were informed that (i) the payoff distributions used for group payment could differ from the examples, and (ii) the group decision would determine binding payoffs at higher stakes.
Unlike in the individual phase, participants did not know the specific distributions in advance.
The discussion lasted at least five minutes and ended with a verbal consensus and a confirming secret-ballot vote; if unanimity was not achieved, payoffs were determined by a random draw.
Finally, participants submitted a last ranking with confidence ratings.

%%%%%%%%%%%%%%%%%
%% Translation %%
%%%%%%%%%%%%%%%%%

\subsection{NormCoRe Translation to AI Agents} \label{results:translation}
Following the NormCoRe translation procedure, replication is treated as a layered translation problem rather than an assumption of equivalence.
Translation decisions are therefore made explicit and justified at each analogy layer.

\textbf{(Step 1) Identify constructs \& study design components:}
The baseline experiment investigates normative preferences for distributive justice principles under a Rawlsian veil of ignorance.
The focal construct is the selection and ranking of four predefined justice principles (P1–P4), measured at both the individual level (repeated rank-orderings with confidence) and the group level (unanimous consensus determining payoffs).
Core design components that operationalize this construct include:
(i) controlled information about income distributions and probabilities,
(ii) payoff-relevant decision making via stochastic assignment to income classes, and (iii) structured group deliberation culminating in a binding collective choice.
All translation choices in the AI agent study are evaluated in relation to their ability to preserve the fairness decision problem while rendering it executable for LLM-based agents.

\textbf{(Step 2) Instantiate analogous layers:}
NormCoRe aligns the human-subject experiment with the AI-agent study through a set of layered analogies.
A complete, parameter-level specification of all mappings is provided in~\cref{tab:normcore_translation_cog_ontho,tab:normcore_translation_interact_interven,tab:normcore_translation_interven,tab:normcore_translation_interven_2} in the Appendix; here we summarize the most salient aspects.

\textit{Cognitive and ontological analogy.}
Human participants---university students deliberating under uncertainty---are mapped to LLM-based AI agents. 
In the baseline study, subjects' cognitive capacities (e.g., language comprehension, memory, and reasoning ability) and ontological properties (e.g., agency, identity, and persistence across interactions) are implicit and embodied.
In the AI agent study, these properties must be made explicit and operationalized.
Accordingly, agents are instantiated with configurable role descriptions approximating the original participant pool, managed memory with explicit character limits to support learning-in-context, and controlled linguistic and stochastic parameters (language choice and temperature).
The complete prompts used to instantiate all agent roles are made available in our repository~\citep{deck2026normcore_software}; consistent with NormCoRe's core principle, they are treated as explicit and auditable translation choices rather than definitive claims about any specific model's normative behavior
This translation preserves the functional role of the subject---forming and revising normative judgments---without asserting equivalence between human subjects and LLM-based agents (\Cref{tab:normcore_translation_cog_ontho,tab:normcore_translation_interact_interven,tab:normcore_translation_interven,tab:normcore_translation_interven_2}).

\textit{Interactional and interventional analogies.}
Human group interaction and experimental procedures are translated into a turn-based orchestration protocol and an agent-integrated workflow.
Free-form discussion is formalized as sequential speaking with equal opportunity, shared discussion history, and a structured consensus mechanism, while the experimenter's role and laboratory environment are translated into structured prompts, computational execution, symbolic payoffs, and explicit randomness controls.
These mappings ensure that deliberation, incentives, and information flow remain comparable across populations while accommodating the technical constraints of AI agents.
Full details of these mappings are documented in~\cref{tab:normcore_translation_cog_ontho,tab:normcore_translation_interact_interven,tab:normcore_translation_interven,tab:normcore_translation_interven_2}.

\textbf{(Step 3) Provide a rationale for translation choices:}
All translation choices are classified; a comprehensive classification and justification for each design decision is provided in~\cref{tab:normcore_translation_cog_ontho,tab:normcore_translation_interact_interven,tab:normcore_translation_interven,tab:normcore_translation_interven_2} in the Appendix.

\textbf{(Step 4) Analyze sensitivity \& robustness:}
NormCoRe requires that translation choices introducing degrees of freedom be either controlled or systematically varied.
Accordingly, the replication incorporates several robustness mechanisms.
First, reproducibility controls (explicit random seeds, bounded discussion rounds, and temperature parameters) enable deterministic reruns and isolate stochastic variation. 
Specifically, we use 33 AI groups to closely approximate the human baseline of 34 groups while allowing the sample to be divisible by three. 
This enables three temperature conditions: $0$ (deterministic), random draws from $[0,1]$, and random draws from $[0,1.5]$, representing increasingly stochastic generation regimes.
Second, sensitivity analyses are embedded directly into the design: agent language (English, Mandarin, Spanish) is varied to assess the stability of normative outcomes.
Third, structural constraints (e.g., validation of floor and range parameters, minimum statement lengths, and capped memory) prevent degenerate or nonsensical trajectories that would confound interpretation.

\subsection{Results of the Replication Study} \label{results:results}
\begin{figure*}
    \centering
    \includegraphics[width=0.85\linewidth]{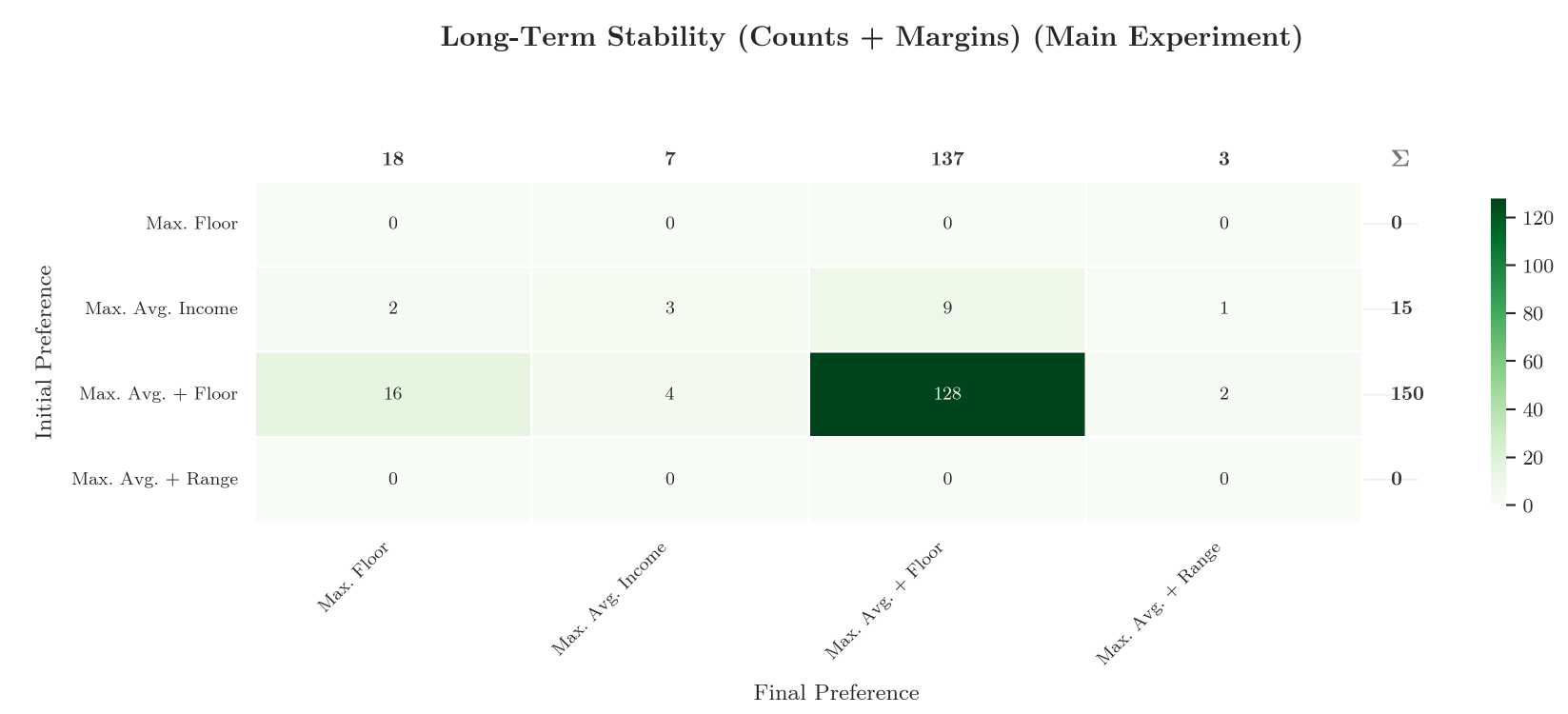}
    \caption{Individual-level preference ranking transitions before and after group deliberation.
    The vertical (horizontal) axis shows initial (final) individual rankings, and cell intensities reflect transition frequencies, highlighting strong convergence toward maximizing average income with a floor constraint.}
    \label{fig:preference_ranking_shifts}
\end{figure*}

We begin by examining \emph{baseline alignment} between the human subject experiment and its MAAI replication.
\Cref{tab:merged_results} reports the distribution of distributive justice principle choices for the human baseline and the aggregated MAAI groups.
Consistent with the original human study, both populations predominantly converge on the principle of maximizing average income subject to a guaranteed minimum for the worst-off.
At the same time, notable differences emerge at the collective level.
MAAI groups exhibit stronger convergence and lower disagreement: 29 of 33 AI groups select this principle, compared to 23 of 34 human groups, with disagreement rates of 9.1\% and 20.6\%, respectively.
These findings indicate that while the dominant fairness preference is aligned across populations, normative outcomes in MAAI are more homogeneous than in human groups.
\Cref{fig:preference_ranking_shifts} complements the aggregate results in \Cref{tab:merged_results} by visualizing how \emph{individual-level preference rankings} change over the course of the experiment.
The figure plots each participant's (human) or agent's (MAAI) \emph{initial individual ranking} (elicited prior to any group interaction) against their \emph{final individual ranking}, reported after group deliberation and the emergence of a collective decision.
Across MAAI experiments, individual agent rankings exhibit substantially stronger convergence toward the eventual \emph{group-level consensus} than observed in human groups.
This pattern indicates reduced intra-group variance and stronger aggregation dynamics in MAAI, helping to explain the higher levels of consensus and lower disagreement rates reported in \Cref{tab:merged_results}.

\begin{figure*}
    \centering
    \includegraphics[width=0.95\linewidth]{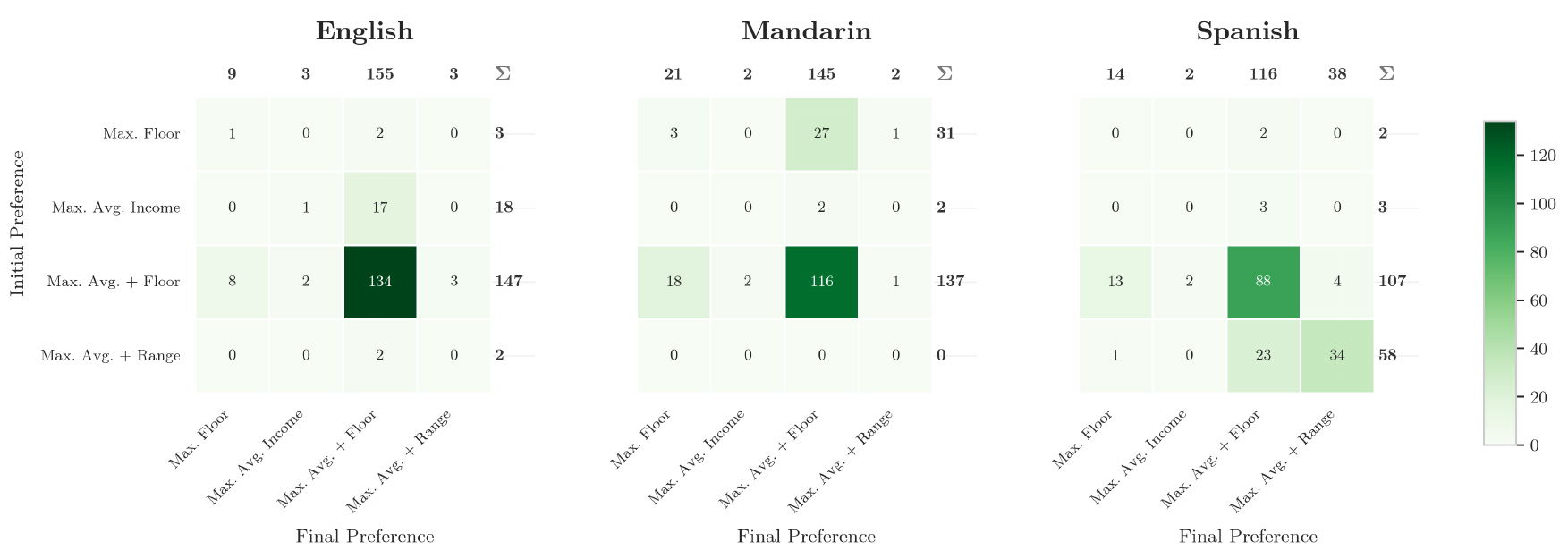}
    \caption{Preference shifts in individual distributive justice rankings before and after group deliberation, stratified by AI agent language.
    Across all languages, individual AI agent preferences converge toward maximizing average income.}
    \label{fig:preference_ranking_shifts_language}
\end{figure*}

Beyond establishing baseline alignment between human and MAAI group-level outcomes, we examine the robustness of these findings to translation-layer design choices.
Sensitivity analyses show that the observed convergence in fairness principles is not invariant, but systematically conditioned by architectural decisions within the AI agent configuration.
At the \emph{cognitive analogy layer}, variation in the underlying foundation model ecosystem (Chinese vs.\ U.S.\ LLMs) leads to pronounced and consistent shifts in distributive justice principle selection, relative to the baseline alignment reported in \Cref{tab:merged_results}.
At the \emph{ontological analogy layer}, sensitivity analyses reveal that the language used to instantiate agent personas systematically conditions individual preference formation and its aggregation during deliberation.
As shown in \Cref{fig:preference_ranking_shifts_language}, agents instantiated in all three languages exhibit convergence toward the same dominant fairness principle observed in the baseline alignment (see \Cref{tab:merged_results}).
However, the degree and trajectory of convergence differ across languages.
Spanish-language agents display more heterogeneous initial individual preferences and retain greater diversity throughout deliberation, whereas English- and Mandarin-language agents begin from more concentrated initial distributions and converge more rapidly toward the group-level consensus.
These language-specific dynamics help explain the variation in consensus rates reported in \Cref{tab:merged_results}, while reinforcing that aggregate alignment with human outcomes can coexist with substantial sensitivity to ontological instantiation choices.

\section{Discussion}
\label{sec:discussion}
The increasing integration of MAAI in both workflow automation and empirical research raises fundamental methodological and normative challenges.
As AI-based agents increasingly deliberate, coordinate, and converge on collective decisions, understanding the emergence and impact of social and ethical norms in such systems becomes critical.
NormCoRe provides a first systematic step toward structuring AI-agent replication studies as an explicit process of translating human subject studies in AI agent studies, acknowledging the fundamental differences between human subjects and AI agents.
Accordingly, we discuss what the experimental results imply for replication-by-translation~(\Cref{discussion:translation}), the epistemic purpose of AI agent studies~(\Cref{discussion:purpose}), and the open challenges and limits of this approach~(\Cref{discussion:challenges}).

\subsection{Why Replication Studies with AI Agents Require Translation} \label{discussion:translation}
The results illustrate both the promise and the limitations of replicating human subject studies with AI agents.
At an aggregate level, MAAI groups converge on the same dominant fairness principle as human groups, suggesting that key normative outcomes can be reproduced under comparable decision structures (\Cref{tab:merged_results}).
At the same time, our experimental results show that MAAI exhibits substantially higher consensus and lower disagreement rates, indicating stronger aggregation dynamics and reduced variance compared to human groups (\Cref{fig:preference_ranking_shifts}).
Crucially, sensitivity analyses show that these outcomes are not invariant.
Varying the foundation model leads to systematic shifts in principle selection (\Cref{tab:merged_results}), and changing the language used to instantiate agent personas affects both convergence and outcome distributions (\Cref{tab:merged_results} and \Cref{fig:preference_ranking_shifts_language}).
These effects demonstrate that seemingly stable normative judgments depend on translation choices that have no direct analogue in the human baseline.
For example, the same persona specification (e.g., a ``university student'') may be instantiated in different prompt languages (English, Mandarin, or Spanish) on foundation models whose pretraining corpora---and thus linguistic and cultural coverage---differ substantially; this introduces variation attributable to translation decisions and model-choice that hardly have a direct analogue in the original human subject study.

While the baseline experiment from\citet{frohlichChoosingJusticeExperimental1992} serves our purpose to instantiate NormCoRe with a suitable and relevant norm, it does not come without limitations and represents only a fraction of what social and ethical norms can entail.
As discussed in the original study, \citet{frohlichChoosingJusticeExperimental1992} only uses a small sample of Polish, Canadian, and US students, which is not nearly representative of the world population, and adopts a strong simplification of norms into four distributive justice principles, which influences the consensus finding process.
Despite these limitations, our findings suggest that similarity in aggregate outcomes is insufficient to establish equivalence between human and AI-based replications.
Instead, replication outcomes are already shaped by design decisions at the cognitive and ontological levels of analogy.
This underscores the need to treat replication with AI agents as a process of explicit translation rather than direct substitution---precisely the role NormCoRe is designed to fulfill.

\subsection{Purpose of AI Agent Studies} \label{discussion:purpose}
Broadening the discourse on AI agent studies, a first open question concerns the epistemic purpose of replication studies with AI agents, e.g., what kind of value such studies provide, and what conclusions should be drawn from them.
One motivation can be curiosity-driven comparison, e.g., observing how AI agents behave when placed in experimental settings originally designed for humans.
However, a purely descriptive comparison risks under-theorizing the implications of observed similarities or differences.
A more substantive motivation is methodological.
The scientific community is currently grappling with a well-documented replication crisis, characterized by systematic failures to reproduce published findings across disciplines~\citep{Baker2016}.
Large-scale efforts, such as the Open Science Collaboration's replication of 100 psychological experiments, revealed replication failures in more than half of the cases, with substantially reduced effect sizes \citep{Open_Science_Collaboration_2015}.
AI-agent replication studies cannot resolve this crisis in a straightforward sense, as they do not constitute replications within the same population.
However, when framed appropriately, they can serve as boundary tests that probe the robustness of theoretical constructs under radically different cognitive and organizational substrates~\citep{Cui2025}.
From this perspective, divergence between human and AI-agent outcomes is not a failure but an informative signal.
This raises a related question: Are AI-agent studies primarily concerned with norms among AI agents themselves, or with downstream outcomes that affect (human) norms?
NormCoRe deliberately accommodates both perspectives.
On the one hand, collective norms emerging within MAAI systems are increasingly consequential in their own right, as such systems are delegated fairness-sensitive decisions in practice.
On the other hand, understanding how and why AI-agent norms differ from human baselines is essential for anticipating societal impacts and governance challenges.

A recurring critique in the literature questions whether it is appropriate to conduct social or behavioral experiments with AI agents at all~\citep{bailCanGenerativeAI2024, xuAISocialScience2024}.
Concerns include sampling bias, the absence of lived experience, and the risk of over-interpreting artificial behavior as psychologically meaningful.
These concerns are valid and underscore the importance of methodological caution.
At the same time, AI agent studies offer distinctive methodological advantages over human subject studies~\citep{mittelstadt2024large}.
Data can be collected at scale, under controlled conditions, and with levels of process transparency that are often unavailable in human-subject research (e.g., by collecting transcripts of thinking models).
Moreover, AI agent replications can explicitly vary dimensions (e.g., language, memory, or interaction protocols) that are difficult or impossible to manipulate cleanly in human experiments.
Generally speaking, to determine the value of MAAI for different applications and the conditions under which MAAI can actually be beneficial, we need to better understand the structure, dynamics, and impact of MAAI systems in the first place. We position NormCoRe as a methodlogical framework to advance this understanding.

\subsection{Open Challenges of AI Agent Studies} \label{discussion:challenges}
Despite their promise, AI-agent replication studies face several unresolved challenges.
A central issue is generalizability.
Empirical findings in AI agent studies are inherently time-bound snapshots: foundation models evolve rapidly, whereas human biology and many social mechanisms remain comparatively stable over time.
As a result, replication outcomes may change as models are updated, retrained, or replaced. 
This temporal instability complicates the accumulation of knowledge and reinforces the need for precise and layered documentation of model versions and experimental configurations.
A related challenge concerns establishing best practices.
There is currently little consensus on how sample sizes should be selected or reported in AI agent studies, particularly when agents can be instantiated cheaply and repeatedly. 
Similarly, while NormCoRe emphasizes explicitation as a core methodological principle, open questions remain regarding the sufficiency of detail.
Moreover, it is unclear how well the behavior of AI agents in an experimental study can be generalized to applied settings.
For example, if AI agents systematically associate certain social traits with gender in an experimental context, this may carry over to applied domains such as hiring or evaluation. 
Emerging interpretability research provides preliminary support for this assumption, suggesting that LLMs internally organize information into relatively stable conceptual representations~\citep{lindsey2025biology}. 
Nonetheless, this line of research remains in its infancy, and strong claims about an AI system's ``psychology''~\citep{hagendorffMachinePsychology2024} would be premature.

Cultural generalizability poses another open question.
Most current AI---and especially LLM---research relies on models trained predominantly on WEIRD (Western, Educated, Industrialized, Rich, and Democratic) data and instantiated with personas reflecting Western norms~\citep{mihalceaWhyAIWEIRD2025}.
A similar tendency can be observed for participant sampling in AI-related human subject studies~\citep{WEIRDFAccTs2023}.
It remains unclear whether non-WEIRD MAAI---or the same systems instantiated with different linguistic and cultural priors---would converge on similar norms or diverge systematically.
Conversely, it is an open empirical question whether some AI agent configurations may align more closely with non-WEIRD human populations than with the original human baselines.
Moreover, hybrid replications, in which one component of a socio-technical system is held constant while another is translated (e.g., human decision-makers interacting with AI agents, or vice versa), blur the boundary between human-subject and AI agent studies and offer additional degrees of freedom~\citep{broska}.
Similarly, one can imagine translating insights from AI agent studies back into human subject research, which raises several questions on the methodology and the interpretation of results.

Lastly, our study has only shown \textit{that} design choices significantly affect normative outcomes, but reveal little about the exact mechanisms underlying these effects.
NormCoRe helps to make influential design factors explicit and supports systematic sensitivity and robustness analyses.
Still, illuminating the causal patterns and technical mechanisms behind observed effects requires complementary approaches such as interpretability, ablation studies, or interdisciplinary analysis (e.g., augmented by linguistic or cultural research), which offer exciting avenues for future work.

Taken together, these challenges point to a broader research agenda.
Key open questions concern the purpose and value of replication studies with AI agents, the generalizability of observed human–AI differences given unavoidable translation and explicitation choices, the generalization of observed norms across applications and cultures, and investigation into the causal patterns driving the observed effects.
NormCoRe does not resolve these questions, but paves the way toward an established methodological foundation, allowing future research to systematically address them.
By advocating for rigorous replication-by-translation, NormCoRe contributes to transparency, interpretability, and cumulative progress in the study of social and ethical norms in MAAI.
\section{Conclusion and Outlook}
\label{sec:conclusion}
As AI agents are increasingly integrated into experimental studies and decision-making processes, methodological rigor in designing these agents becomes critical---especially when they are subject to social and ethical norms. 
To judge whether and under which conditions the integration of AI agents is actually beneficial, we need to better understand their structure, dynamics, and impact.
We introduce NormCoRe as a methodological framework for studying social and ethical norms---such as fairness---in Multi-agent AI (MAAI) setups through the rigorous replication-by-translation of human subject studies in AI agent environments.
By accounting for the fundamental differences between human subjects and AI agents and conceiving replication as a layered process of analogous translation, NormCoRe systemizes the design choices shaping normative judgments and outcomes in MAAI systems.
Our experimental study, which instantiates the NormCoRe method, demonstrates that fairness judgments in MAAI are sensitive to the choice of the foundation model and the language used to instantiate agent personas. The results reveal significant differences between human subjects in their normative judgments and underscore the importance of well-documented design choices for the examined AI agents.
Also, our study indicates that AI agents can converge on fairness principles similar to those favored by human groups, but do so with higher homogeneity, highlighting the importance of accounting for the differences between human subjects and AI agents.

Looking ahead, our work provides a blueprint for investigating normative dynamics in MAAI beyond distributive justice, which opens several avenues for future research and applications. 
First, while our study offers early insights into fairness principles in MAAI, NormCoRe should be applied to other social and ethical norms (e.g., transparency, reciprocity, or trust) to gain a deeper understanding of the mechanics and potential risks associated with MAAI systems.
Second, applying NormCoRe to hybrid settings that combine human subjects and AI agents may illuminate novel dynamics emerging in MAAI systems (e.g., power asymmetries and coordination effects).
Lastly, beyond empirical studies NormCore helps to guide, reflect, and document design choices whenever AI agents are used to automate or support tasks formerly carried out by (groups of) humans. Future work should establish best practices and standards for certain design choices to facilitate longitudinal and cross-cultural studies, which are critically needed as models, data, and deployment contexts evolve rapidly.
While the risks and benefits of increased adoption of AI agents are disputed, our work invites the community to view AI agents not merely as something that needs to be aligned with human norms and values, but as an evolving technology demanding scrutiny and systematic evaluation.
Understanding and governing these agents requires methods that ensure scientific rigor and accountability for design decisions.
Ultimately, only through such foundational work can we ensure to uphold social and ethical norms in an increasingly automated world.

\section{Generative AI Disclosure Statement}
For writing, we used ChatGPT, Grammarly and DeepL to improve grammar and fluency. Moreover, to develop the software artifact used in the experimental study, we used the coding tools Claude Code, Codex CLI, and Gemini CLI.

\bibliographystyle{ACM-Reference-Format}
\bibliography{refs}


\appendix
\appendix
\section*{Appendix}

%%%%%%%%%%%%%%%%%%%%%%%%%%%%%%%%%%%%%%%%%%%%%%%%%%%%%%%%%%%
%%%%%%%%%%%%%%%%%%%%%%%%%%%%%%%%%%%%%%%%%%%%%%%%%%%%%%%%%%%
%%%%%%%%%%%%%%%%%%%%%%%%%%%%%%%%%%%%%%%%%%%%%%%%%%%%%%%%%%%
\begin{table*}[t]
\footnotesize
\centering
\caption{NormCoRe translation table for cognitive and onthological analogy}
\label{tab:normcore_translation_cog_ontho}

\adjustbox{angle=90, max width=\textheight}{
\begin{tabular}{p{1.4cm} p{2.5cm} p{1.3cm} p{2cm} p{4cm} p{5.9cm}}
\toprule
\textit{Step 1} &
\textit{Step 1} &
\textit{Step 2} &
\textit{Step 3} &
\textit{Step 2} &
\textit{Step 3} \\
\toprule
\textbf{Aspect} &
\textbf{Human Subject Study} &
\textbf{Analogy Layer} &
\textbf{Translation Choice} &
\textbf{AI Agent Study Instantiation} &
\textbf{Rationale} \\
\midrule

Participant Sampling &
Convenience sample of university students ($n=34$ human groups) &
Cognitive &
Explicitation (Incremental) &
Selection of sufficiently capable foundation models with 33 AI groups approximating human sample size.
Basline alignment: \textit{gemini-2.5-pro}, \textit{gemini-2.5-flash}, and \textit{gemini-2.5-flash-lite}.
Translation sensitivity (cognitive): \textit{DeepSeek-V3.2 Experimental}, \textit{ChatGLM-4.5-Air}, and \textit{Qwen3-Omni 30B A3B}; \textit{gpt-oss-
120B high}, \textit{Grok Code Fast 1}, and \textit{Grok 4-Fast}.
Translation sensitivity (ontological) similar to basline alignment.
&
Human sampling implicitly ensures sufficient cognitive and linguistic capacity to understand the task; in AI agent studies, this assumption must be made explicit by selecting foundation models with adequate reasoning, language comprehension, and instruction-following capabilities.
While AI agent studies could easily scale to much larger samples, NormCoRe approximates the original sample size to preserve comparability of group-level outcomes and avoid introducing asymmetries driven by computational scalability.
In other contexts, larger samples may be appropriate, e.g., scaling behavior or statistical power is itself the object of study.\\

Statement Validation &
Unconstrained natural speech &
Cognitive &
Explicitation (Incremental) &
Minimum statement length with retry mechanism &
Basic output constraints prevent degenerate or empty responses, thereby improving the robustness of the experimental artifact. \\

\midrule

Participant Type &
Human subjects (university students) &
Ontological &
Literal Translation (Direct) &
LLM-based agents &
Fundamental premise of the experiment: normative judgments are produced by deliberating agents embedded in groups. \\

Participant Personality &
Natural human variation &
Ontological &
Explicitation (Incremental) &
Configurable Role Description &
Human personality variation is implicit in the original study; agents are explicitly instructed to behave like college students to approximate the participant pool. \\

Participant Memory &
Human biological memory with natural limitations &
Ontological &
Explicitation (Constructive) &
Agent-managed memory with character limits &
Human memory is implicit and embodied; for AI agents it must be formalized to enable learning-in-context and consistency across deliberation rounds. \\

Participant Language &
Primarly English (flawed Polish) &
Ontological &
Explicitation (Quasirandom) &
Configurable prompt and agent language (English, Mandarin, Spanish) &
Language is implicit and fixed in the human study; explicit variation enables testing the sensitivity of normative outcomes to linguistic framing. \\

Temperature Control &
\textit{Not applicable} &
Ontological &
Explicitation (Comprehensive) &
Configurable LLM temperature parameter for response randomness. &
No human analogue exists; response stochasticity is explicitly parameterized to ensure reproducibility and controlled variation.\\

\bottomrule
\end{tabular}
}
\end{table*}

\begin{table*}[t]
\footnotesize
\centering
\caption{NormCoRe translation table for interactional and interventional analogy}
\label{tab:normcore_translation_interact_interven}
\adjustbox{angle=90, max width=\textheight}{
\begin{tabular}{p{1.4cm} p{2.5cm} p{1.3cm} p{2cm} p{4cm} p{5.1cm}}
\toprule
\textit{Step 1} &
\textit{Step 1} &
\textit{Step 2} &
\textit{Step 3} &
\textit{Step 2} &
\textit{Step 3} \\
\toprule
\textbf{Aspect} &
\textbf{Human Subject Study} &
\textbf{Analogy Layer} &
\textbf{Translation Choice} &
\textbf{AI Agent Study Instantiation} &
\textbf{Rationale} \\
\midrule
\midrule

Experiment Duration &
Unlimited Time &
Inter-actional &
Explicitation (Constructive) &
Fixed maximum number of rounds (default: 10), with configurable upper bounds &
LLM agents cannot speak simultaneously; therefore, the number of interaction rounds must be bounded to control computational cost. \\

Question Asking &
Natural Questions to experimenter &
Inter-actional &
Explicitation (Incremental) &
Not supported in dynamic orchestration &
Supporting ad hoc questions would substantially increase system complexity, and it cannot be guaranteed that responding AI agents would provide accurate or authoritative information. \\

Discussion Format &
Free-form group discussion &
Inter-actional &
Explicitation (Constructive) &
Turn-based sequential discussion with equal speaking opportunities &
Human conversational norms must be made explicit in AI systems; turn-based interaction ensures procedural fairness and comparability across agents, independent of response latency. \\

\midrule

Prompt Structure &
Identity &
Inter-ventional &
Explicitation (Constructive) &
At each interaction, AI agents receive an instruction prompt containing their name, role description, bank balance, current experimental phase, and memory state (in Phase~2, including the names of other participants and the discussion history), each separated by line breaks. The input prompt separately specifies the agent’s current task. &
High-level agent identity is explicitly separated from task-specific instructions to distinguish stable background characteristics from situational decision demands and to hierarchically structure information from global to local context. \\
\bottomrule
\end{tabular}
}
\end{table*}

%%%%%%%%%%%%%%%%%%%%%%%%
% TABLE INTERVENTION 2 %
%%%%%%%%%%%%%%%%%%%%%%%%

\begin{table*}[t]
\footnotesize
\centering
\caption{NormCoRe translation table for interventional analogy}
\label{tab:normcore_translation_interven}
\adjustbox{angle=90, max width=\textheight}{
\begin{tabular}{p{1.4cm} p{2.5cm} p{1.3cm} p{2cm} p{2.5cm} p{6.6cm}}
\toprule
\textit{Step 1} &
\textit{Step 1} &
\textit{Step 2} &
\textit{Step 3} &
\textit{Step 2} &
\textit{Step 3} \\
\toprule
\textbf{Aspect} &
\textbf{Human Subject Study} &
\textbf{Analogy Layer} &
\textbf{Translation Choice} &
\textbf{AI Agent Study Instantiation} &
\textbf{Rationale} \\
\midrule
\midrule

Environment &
Laboratory setting with physical presence &
Inter-ventional &
Literal translation (Direct) &
Computational / Data environment &
The physical laboratory environment is directly translated into a computational setting, preserving the experimental context while adapting it to non-embodied agents. \\

Payment Mechanism &
Real monetary payment based on chosen principle &
Inter-ventional &
Explicitation (Constructive) &
Symbolic payoff with explicit bank balance updates &
Because AI agents cannot receive physical monetary payments, payoffs are explicitly represented symbolically via balance updates, introducing a new representational structure. \\

Probability Calculations &
Original probability values &
Inter-ventional &
Explicitation (Incremental) &
Recalculated probabilities with explicit assumptions for Situation~C &
Minor assumptions are introduced to resolve underspecification in the original probability structure while preserving the intended payoff logic. \\

Distribution Presentation &
Tabular presentation to subjects &
Inter-ventional &
Literal translation (Direct) &
Structured prompt-based presentation of distribution details &
The informational content of the original tabular presentation is preserved and directly translated into a text-based prompt format. \\

Payoff Presentation &
Single realized payoff and counterfactual outcomes &
Inter-ventional &
Explicitation (Comprehensive) &
Explicit presentation of realized and counterfactual payoffs for all principles &
The payoff presentation is substantially expanded to fully explicate counterfactual information required for consistent interpretation by AI agents. \\

Floor / Range Constraint Options &
Range of permissible floor and range constraint values &
Inter-ventional &
Explicitation (Comprehensive) &
Validated and constrained floor and range parameters with error handling &
Implicit plausibility constraints in the human study are fully formalized through validation rules and corrective feedback. \\

Experimenter Instructions &
Verbal instructions from human experimenter &
Inter-ventional &
Explicitation (Constructive) &
Structured JSON-based prompts across all experimental stages &
Experimenter instructions are formalized into structured prompts to ensure consistency and reproducibility without introducing an artificial authority agent. \\

Distributions in Application Rounds &
Four predefined situations (A--D) &
Inter-ventional &
Explicitation (Incremental) &
Original distributions adopted with minor probability assumptions &
Some probability values are not explicitly specified in the original study and are inferred under minimal assumptions to render the distributions computationally executable. \\

\bottomrule
\end{tabular}
}
\end{table*}

%%%%%%%%%%%%%%%%%%%%%%%%
% TABLE INTERVENTION 3 %
%%%%%%%%%%%%%%%%%%%%%%%%

\begin{table*}[t]
\footnotesize
\centering
\caption{NormCoRe translation table for interventional analogy}
\label{tab:normcore_translation_interven_2}
\adjustbox{angle=90, max width=\textheight}{
\begin{tabular}{p{1.4cm} p{2.6cm} p{1.3cm} p{2.2cm} p{2.8cm} p{6.2cm}}
\toprule
\textit{Step 1} &
\textit{Step 1} &
\textit{Step 2} &
\textit{Step 3} &
\textit{Step 2} &
\textit{Step 3} \\
\toprule
\textbf{Aspect} &
\textbf{Human Subject Study} &
\textbf{Analogy Layer} &
\textbf{Translation Choice} &
\textbf{AI Agent Study Instantiation} &
\textbf{Rationale} \\
\midrule

Comprehen-sion Test &
Comprehension test to verify participant understanding &
Inter-ventional &
Explicitation (Incremental) &
Comprehension test omitted &
Human comprehension checks ensure baseline task understanding; for AI agents, task comprehension is enforced through prompt design, making an explicit test redundant and unnecessarily complex. \\

Speaking Order &
Emergent conversational order &
Inter-ventional &
Explicitation (Incremental) &
Randomized speaking order with constraints &
Randomization mitigates systematic ordering effects (e.g., first- or last-mover advantages) while preserving the deliberative structure of the discussion. \\

Discussion History Management &
Natural human memory recall &
Inter-ventional &
Explicitation (Constructive) &
Explicit provision of shared discussion history with bounded context length &
Human memory is implicit and selective; in AI agents, shared memory must be explicitly provided and bounded to prevent context overflow while maintaining deliberative continuity. \\

Consensus Process &
Verbal consensus and secret-ballot confirmation &
Inter-ventional &
Explicitation (Constructive) &
Formalized two-stage secret ballot with validation &
Consensus formation is operationalized explicitly to ensure unambiguous termination and verifiable agreement in the absence of non-verbal cues. \\

Vote Initiation &
Implicitly emerges during discussion &
Inter-ventional &
Explicitation (Constructive) &
Explicit per-round vote-initiation query with confirmation protocol &
Human cues for vote initiation have no direct analogue; an explicit protocol ensures consistent triggering of the voting phase across models with varying capabilities. \\

Novel Principles &
Participants could propose new principles &
Inter-ventional &
Explicitation (Incremental) &
Proposal of novel principles disabled &
Allowing novel principles would substantially expand the decision space; this constraint mirrors empirical findings that no novel principles emerged in the original study. \\

Internal Thinking &
Implicit cognitive deliberation &
Inter-ventional &
Explicitation (Constructive) &
Private strategic assessment prompt per round &
Explicit internal reasoning prompts support structured deliberation by AI agents and compensate for the absence of implicit cognitive processes. \\

Payoff Calculation &
Higher but unspecified stakes in group phase &
Inter-ventional &
Explicitation (Incremental) &
Original distributions scaled by random factor (2--6) &
Because the original study specifies higher stakes without defining distributions, proportional scaling preserves the payoff logic while covering a plausible outcome range. \\

\bottomrule
\end{tabular}
}
\end{table*}

\end{document}